%% file: elasticsolver.tex
\documentclass[10pt, conference, compsocconf]{IEEEtran}
\usepackage{booktabs}
\usepackage{paralist}
\usepackage{graphicx}

\title{Elastic Solver: Balancing Solution Time and Energy Consumption}
\author{\IEEEauthorblockN{Barry Hurley, Deepak Mehta, and Barry O'Sullivan}
\IEEEauthorblockA{Insight Centre for Data Analytics\\
University College Cork, Ireland\\
\{barry.hurley$|$deepak.mehta$|$barry.osullivan\}@insight-centre.org}
}

\begin{document}

\maketitle

\begin{abstract}

Combinatorial decision problems arise in many different domains such as scheduling, routing, packing, bioinformatics, and many more.
Unfortunately,  most of these problems are NP-complete. 
Despite recent advances in developing scalable solvers,
 there are still many problems which are often very hard to solve.
Typically the most advanced solvers include elements which are stochastic in nature.
If a same instance is solved many times using different seeds then depending on the inherent characteristics of a problem instance and the solver, one can observe a highly-variant distribution of times spanning multiple orders of magnitude. 
Therefore, to solve a problem instance efficiently it is often useful to solve the same instance in parallel with different seeds. 
With the proliferation of cloud computing, it is natural to think about an elastic solver which can scale up by launching searches in parallel on thousands of machines (or cores).  
However, this could result in consuming a lot of energy.
Moreover, not every instance would require thousands of machines.
The challenge is to resolve the tradeoff between solution time and energy consumption optimally for a given problem instance.
We analyse the impact of the number of machines (or cores) on not only solution time but also on energy consumption.
We highlight that although solution time always drops as the number of machines increases, 
the relation between the number of machines and energy consumption is more complicated. 
In many cases, the optimal energy consumption may be achieved by a middle ground, we analyse this relationship in detail.
The tradeoff between the solution time and energy consumption is studied further, showing that the energy consumption of a solver can be reduced drastically if we increase the  solution time marginally.
We also develop a prediction model using machine learning, demonstrating that such insights can be exploited to achieve faster solutions times in a more energy efficient manor.
\end{abstract}


\section{Introduction}

Energy consumption for cloud providers and data centres is a growing concern, being one of the largest consumers of electricity.
If current practices for data consumption and processing continue, by 2040 the entire global energy supply will be consumed by large scale data centres.
Recently, practitioners in the areas of Constraint Satisfaction (CSP)~\cite{CPHandbook}, Boolean satisfiability (SAT)~\cite{HandbookOfSAT2009}, Integer Programming (IP)~\cite{Wolsey1998}, and numerous other combinatorial search frameworks have turned to the cloud to solve larger and more challenging combinatorial problems efficiently.
Many industrial solvers such as IBM ILOG CPLEX and Gurobi already exploit the elasticity of the cloud.  These solvers can run on  many machines in parallel to solve difficult combinatorial problems.
The traditional view of parallel computing has focused on minimising execution time in which case one might simply launch the solver on all the available machines.
An issue arises in that one does not know a priori the optimal number of machines to be used in parallel, nor has the energy consumption of such a decision been considered.
In the context of solving combinatorial problems in the cloud, solution time alone cannot be viewed as a single objective.
Instead, one needs to assess the tradeoff in solution time against energy consumption. In our context, the total energy consumption is approximated by the solution time multiplied by the number of searches done in parallel (number of cores).

In general, solving combinatorial search problems is an $\mathcal{NP}$-complete task, typically solved using a combination of search and inference to prune the search space.
Choices for parameters such as the search heuristics, restarting policy, and even random seed can affect the size of the search space and subsequently the time it takes to find a solution~\cite{DBLP:series/faia/GomesS09}. 
Variable and value selection heuristics have elements which are stochastic in nature, so the slightest difference over repeated runs can magnify the performance variations~\cite{DBLP:conf/ijcai/0001O15}.
Thus, modern combinatorial search solvers often exhibit a very high variation in solution time.
Such variations can be modelled by heavy- or fat-tailed distributions~\cite{Gomes:2000wp}.
Intuitively, these model a non-negligible probability of a solver taking exponential time.
However, the runtime distributions can be exploited, either by randomised restarting~\cite{DBLP:series/faia/GomesS09}, or parallelisation~\cite{Hogg:1994tf}.
An instance for which the runtime is variable may be solved more-effectively if several searches are performed in parallel using  different seeds, with search terminating as soon one finds a solution.

In this respect, we exploit the runtime distribution through parallel searches and study its impact, not only on solution time but also on energy consumption as the number of CPU-cores (or machines) is increased.
We show that the relationship between the number of cores and the total energy consumption is not a simple linear relationship.
The natural thought is to assume that as the number of machines is increased, the energy increases correspondingly. In fact, in many cases the minimal energy consumption may be achieved by a using a larger number of machines, with the increased likelihood of finding a solution faster meaning the search can be terminated sooner across all machines, resulting in a reduced energy consumption overall.
Secondly, we analyse the trade-off between the solution time and the total energy consumption.
We will motivate the need of a multi-objective optimisation problem
for  deciding the number of (virtual) machines offered by cloud providers 
in order to strike a balance between solution time and energy consumption.
Finally, we demonstrate that it is possible to use machine learning to predict the number of machines (or number cores) that are required to meet a desired level of balance between energy consumption and solution time.

\section{Time vs Energy}

This section analyses the behaviour of the solution time and the total energy with respect to the number of cores.
Without lack of generality we assume each physical machine is associated with  one CPU-core. 
%
The benchmark set comprises of 1676 industrial instances of combinatorial problems from 9 years of the SAT Competitions, Races, and Challenges between 2002 and 2011~\cite{satcomp}.
Each instance was run using MiniSat~2.0~\cite{MiniSAT} as the solver with 100 different seeds, a timeout of 1 hour CPU-time for each run, and a limit of 2GB RAM. Performance data was collected on a cluster of Intel Xeon E5430 Processors (2.66GHz) running CentOS 6. A total of 315~weeks of CPU-time was consumed to accumulate this performance data.
Instances which were not solved within the time limit across any run, or were solved in under 1 second across every run are excluded, leaving a total of 902 challenging industrial instances.

\begin{figure}[h]
\centering
\includegraphics[width=\columnwidth]{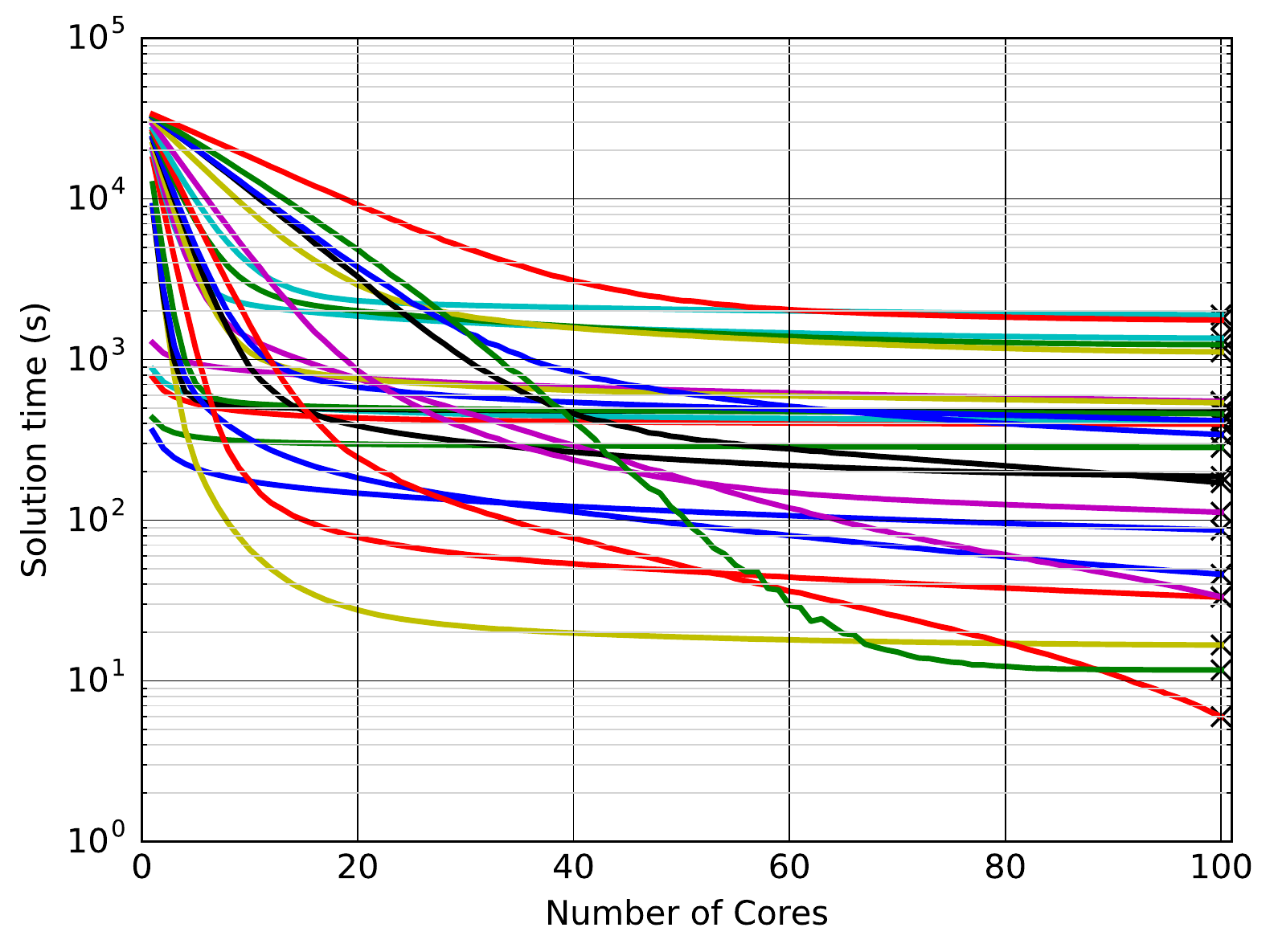}
\caption{Illustration of solution time versus the number of cores for some sample instances. 
}
\label{fig:soltimevscores}
\end{figure}

If the distribution of the runtime is known, then order statistics may be used to model the expected time of running $k$ parallel searches. However, since many of the instances considered did not fit any well known distribution with high-confidence, we sample from the empirical distributions, taking the minimum of $k$ samples. This is repeated by iterating 100,000 times to get an expected time.
Figure~\ref{fig:soltimevscores} illustrates how the expected solution time changes as the number of cores (number of parallel searches) increases.
Only a sample of the most challenging instance are presented, but they are representative of the complete data set.
Naturally, solving the same instance many times in parallel by using more  cores reduces the solution time. It is interesting to see that in some cases multiple orders of magnitude speedup can be achieved by only a handful of additional cores.
In general the solution time for any given problem instance  is non-increasing with respect to  the number of cores.

\begin{figure}[h]
\centering
\includegraphics[width=\columnwidth]{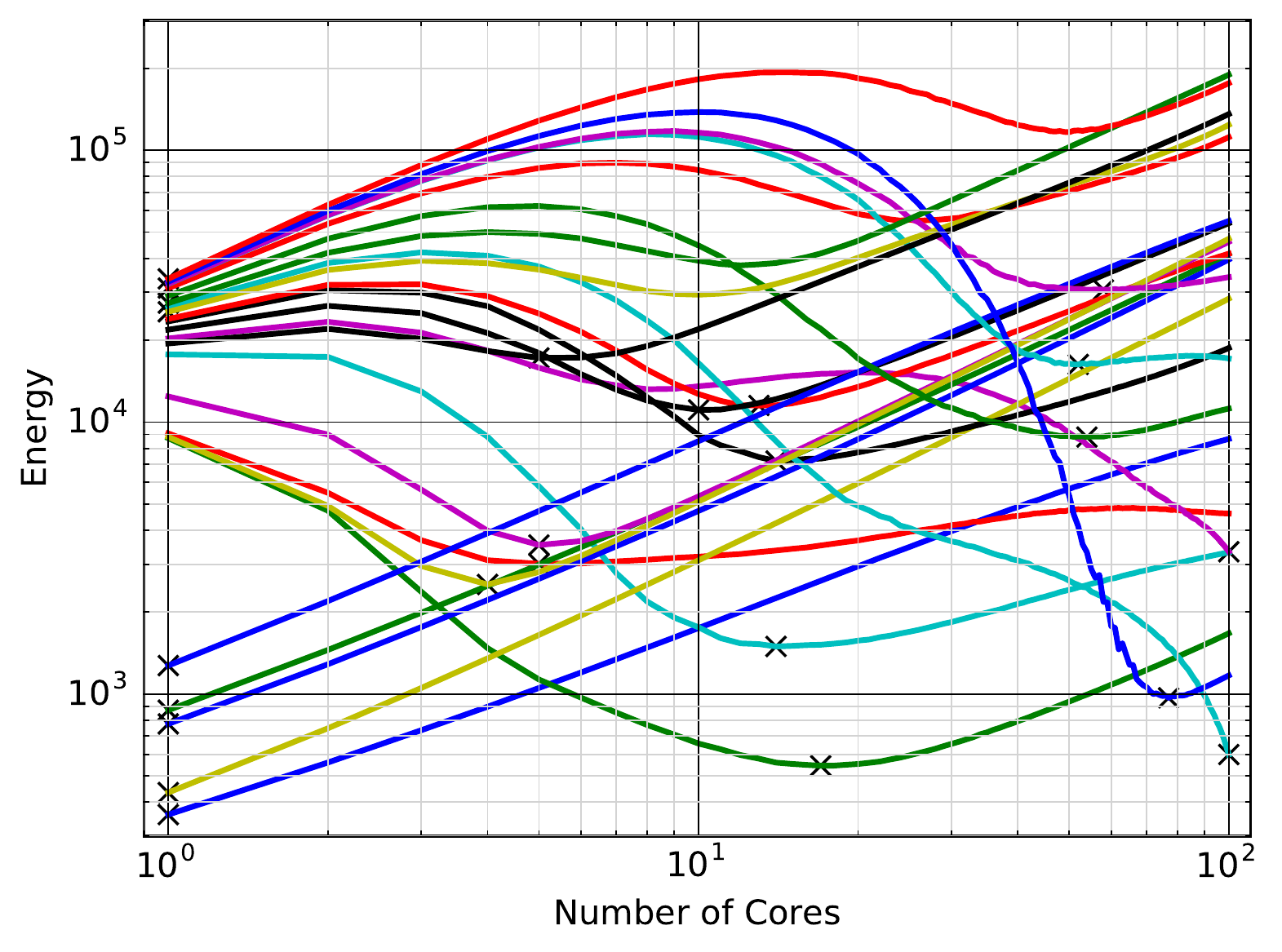}
\caption{Illustration of the total energy versus the number of cores in log-scale for some sample instances. The minimal energy point is marked for each instance.}
\label{fig:energyvscores}
\end{figure}

Figure~\ref{fig:energyvscores} illustrates the energy consumed with respect to  the number of cores for the same set of instances as used in Figure~\ref{fig:soltimevscores}.
	Let $s_k$ denotes the expected solution time using $k$ cores.
The energy consumed using $k$ cores is going to be proportional to the expected solution time with respect to $k$ cores multiplied by $k$. 
Although the expected solution time is non-decreasing with respect to number of cores, the product of the number of cores and the expected time results in a number of interesting profiles.
Sometimes the energy cost initially decreases as the number of cores increases, reaches a minima, and steadily climbs again.
In other cases, the energy cost initially increases with respect to the number of cores and thereafter it declines as the number of cores increases further. 
Other interesting profiles are also visible in the figure.
Evidently, there is no consistent behaviour between instances which achieves the minimal energy cost.
The relationship between the total energy consumed with respect to the number of cores is more complicated 
as evident by a variety of behaviours shown in the figure.

The total energy consumed for solving an instance depends on the run-time distribution.
For example, if in certain cases the runtime distribution is uniform, 
then  the minimal energy cost is achieved by sticking to a single core, adding any more only serves to increase the energy cost.
This is because the expected solution time for any given number of cores does not change when the distribution is uniform.
In contrast, if the distribution is heavy-tailed  and if the expected solution time using 100 cores is  100 times less than the   running time using 1 core for a given instance then  the most energy efficient manor is by running it on 100 cores.
Additionally, a middle grounds also exist, where the most energy efficient solution is somewhere between 1~and~100 cores.
Thus, the energy consumed is minimal using $k$ cores if 
$s_k \times k$ is less than $s_k' \times k'$ for any $k' \geq 1$.

\begin{figure}[h]
\centering
\includegraphics[width=\columnwidth]{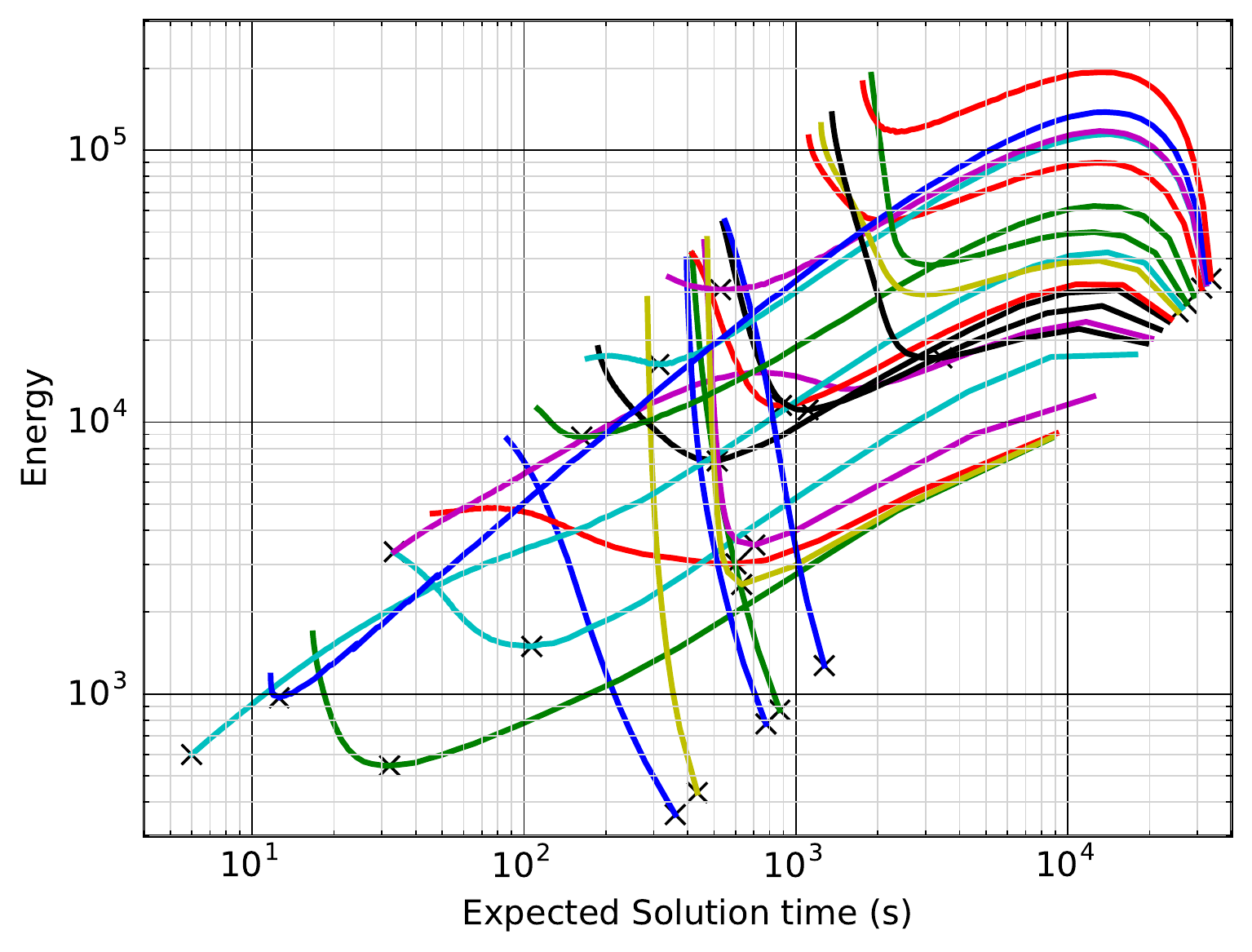}
\caption{Illustration of solution time versus total energy in log-scale for some sample instances. The plus symbol marks the minimal energy for each instance.}
\label{fig:energyvssoltime}
\end{figure}

Figure~\ref{fig:energyvssoltime} plots the trade-off between the expected solution time and the energy consumption for a sample of instances.
It  shows that  the energy consumption curve with respect to solution time can be significantly different for different problem instances.
The  benefit in terms of energy consumption  from independent $k$ parallel searches
is determined by the nature of the full distribution of runtimes.
We  remark that if the expected solution time  is minimum using $k$ cores then the total energy curve would be linear with respect to the number of cores beyond the point $k$. 
In other words, the total energy required by $k'$ where $k' > k$
would always be more than that required by $k$ cores.
Thus, if the expected time stops improving beyond a given number of cores $k$, then any solution obtained by using $k'$ cores
where $k' > k$ would not be part of the pareto-frontier. Consequently it will be dominated by at least one solution obtained using $k''$ cores where $k'' \leq k$.

\begin{figure}[h]
\centering
\includegraphics[width=0.9\columnwidth]{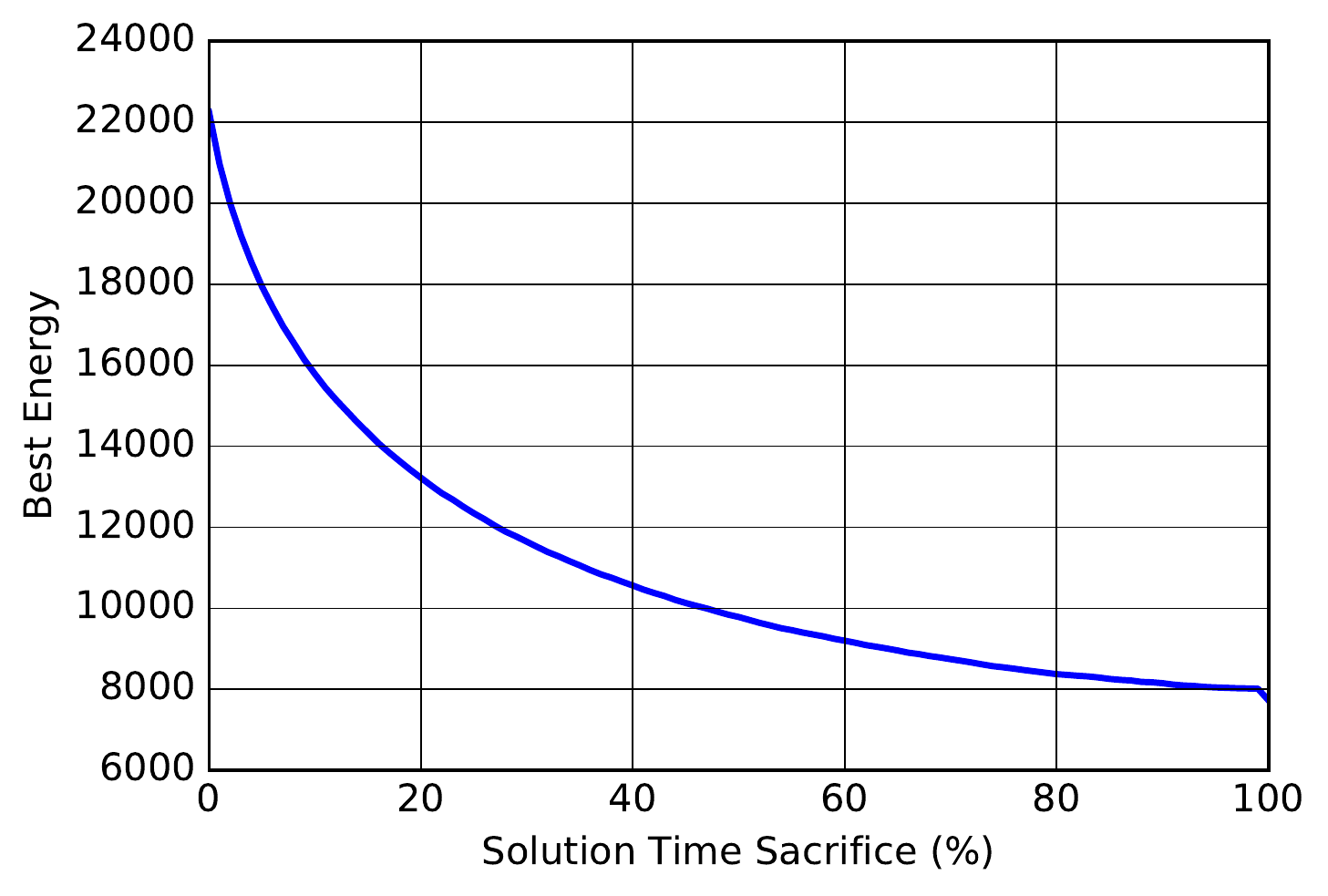}
\caption{Illustration of trade off between solution time and the best energy achievable over all instances.}
\label{fig:soltimevsbestenergy}
\end{figure}

Figure~\ref{fig:soltimevsbestenergy}  presents the trade-off between  the expected solution time and energy consumption, aggregated over all instances. 
The figure depicts  that on average by increasing the solution time by just $10\%$, the energy consumption can be reduced by $20\%$, and by increasing the solution time by $20\%$, the total energy can be reduced by $40\%$. 
Thus,  depending on the preferred bound on the expected solution time, it might be possible to select a number of cores  that minimises the energy consumption on per instance level.
More precisely, the objective would be to predict a number of cores that can minimise the energy consumption and solve  a given problem instance within a given target solution time.

\section{Predicting the Optimal Number of Cores}
\label{sec:ml}

The previous sections have presented evidence that there is no consistent number of cores to run in order to achieve the desired level of balance between energy consumption and solution time. 
This section demonstrates that a machine learning algorithm can be built to exploit this knowledge and make intelligent decisions on an instance specific basis.
In particular, we will develop a model for predicting the number of cores for minimising energy consumption.

To develop a prediction model, the first task is to associate a set of features with each problem instance.
We employ the state of the art collection of 138 features~\cite{SATfeatures}, which have been proved highly-effective in areas of runtime prediction~\cite{Hutter:2014ci,DBLP:conf/ijcai/0001O15} and solver portfolios~\cite{satzilla:jair08,isac:ecai2010}.
Random forest regression is used as the machine learning model, with default parameters except for setting the number of estimator trees to 100.
This model has been shown to be highly effective, robust, and is capable of modelling highly non-linear relations.
The model is built using stratified 10-fold randomised cross-validation. This splits the dataset into 10 equally sized folds with an even distribution of the label in each.
One fold is set aside for testing with the remaining folds used to train the model. This is repeated with each fold taking a turn as the test set.
The goal is to predict the optimal number of cores which minimises overall energy consumption.

Table~\ref{tbl:parallelenergysummary} summarises the comparison between the intelligent machine learning model to various  baseline policies.
Results are sorted by success rate first and then by solution time.
The success rate shows the expected percentage of the jobs to produce a valid result within the specified time limit of 1 hour.
The solution time shows the expect time in which a solution would be found and returned to the user.
As a proxy for the total energy consumed, we use the cumulative CPU-time across all cores.
More sophisticated energy functions may also be employed, but the one used here serves serves to be intuitive.

The first set of baseline policies consider a static approach where the instance is always run on a fixed number of $k$ cores.
Two other baselines correspond to the \emph{virtual best} (VB) energy policy, and the \emph{virtual best solved} policy.
These respectively correspond to an oracle choosing, for each instance, the number of cores leading to
\begin{inparaenum}[i)]
\item the overall minimal energy cost, and
\item the highest expected success rate with minimal energy cost.
\end{inparaenum}

\begin{table}[htb]
\centering
\caption{Evaluation of various  policies.}
\label{tbl:parallelenergysummary}
\input{tblparallelenergysummary-meanpar10.tex}
\end{table}

Firstly, as would be expected, the fixed policy of 1 core is the worst in terms of both success rate and solution time.
Interestingly, the virtual best energy policy, as well as having a lower energy consumption has a slightly better success rate and lower solution time than the single core policy.
As the number of cores in the fixed policies increases, both the success rate and solution time improve, but the overall total energy increases.
Naturally, the largest policy, where all 100 cores are used in parallel provides the highest success rate and best solution time but its energy cost is wasteful.

Most importantly, the machine learning model which predicts the number of cores to be run for each instance can provide a success rate of almost 99\% and a solution time very close to the VB Solved policy.
Interestingly, its energy consumption is much better than that of the VB Solved; by sacrificing a success rate of 1\%, it reduces the total energy usage by 29\%.

\section{Conclusions}
In this paper we have proposed an elastic solver that can balance the solution time and energy consumption.
The solver can scale up in the cloud setting by predicting the number of cores required to strike the balance between the two criterion.
We have studied the behaviour of  the energy consumed by the solver for many real-world industrial instances when different number of cores are used.
Despite the non-trival relationship between solution time and energy, the prediction model is highly effective at predicting the optimal number of cores which will minimise the overall energy consumption.

\bibliographystyle{IEEEtran}
\bibliography{elasticsolver.bib}

\end{document}

%% file: tblparallelenergysummary-meanpar10.tex
\begin{tabular}[c]{rlrrr} \toprule
     & & & \multicolumn{1}{c}{Total} & \multicolumn{1}{c}{Solution} \\
     & \multicolumn{1}{c}{Policy} & \multicolumn{1}{c}{Success\%} & \multicolumn{1}{c}{Energy} & \multicolumn{1}{c}{Time (s)} \\
     \midrule
 1 & Fixed 100 cores & 100.0\% & 23227 & 232.3 \\
 2 & VB Solved & 100.0\% & 7727 & 362.4 \\
 3 & ML Prediction & 98.7\% & 5494 & 366.8 \\
 4 & Fixed 8 cores & 96.4\% & 3228 & 403.5 \\
 5 & Fixed 4 cores & 94.9\% & 1883 & 470.9 \\
 6 & Fixed 2 cores & 93.0\% & 1107 & 553.3 \\
 7 & VB Energy & 92.6\% & 583 & 556.9 \\
 8 & Fixed 1 core & 90.5\% & 654 & 654.2 \\
\bottomrule
\end{tabular}